\documentclass[english]{cccconf}
\usepackage[comma,numbers,square,sort&compress]{natbib}
\usepackage{epstopdf}
\usepackage{enumerate}
\usepackage{tabularx}
\usepackage{amsmath}
\usepackage{amssymb}
\usepackage{amsfonts}
\usepackage{algorithm}
\usepackage{algpseudocode}
\usepackage{amsthm}
\usepackage{graphicx}
\usepackage{epsfig}
\usepackage{setspace}
\usepackage{tensor}

\graphicspath{{figures/}}
\DeclareGraphicsExtensions{.eps}

\theoremstyle{definition}

\newcommand*\diff{\mathop{}\!\mathrm{d}}

\newcommand\T{{\hspace{-0pt}\intercal}}

\DeclareMathOperator{\sat}{sat}

\DeclareMathOperator*{\argmin}{arg\,min}

\begin{document}

\title{On Model Adaptation for Sensorimotor Control of Robots}

\author{David Navarro-Alarcon\aref{polyu}, Andrea Cherubini\aref{lirmm} and Xiang Li\aref{cuhk}}

\affiliation[polyu]{The Hong Kong Polytechnic University, Hung Hom, Kowloon, Hong Kong \email{dnavar@polyu.edu.hk}}
\affiliation[lirmm]{University of Montpellier / LIRMM, Montpellier, France \email{andrea.cherubini@lirmm.fr} }
\affiliation[cuhk]{The Chinese University of Hong Kong, Shatin, New Territories, Hong Kong \email{xli@mae.cuhk.edu.hk}}

\maketitle

\begin{abstract}
In this article, we address the problem of computing adaptive sensorimotor models that can be used for guiding the motion of robotic systems with uncertain action-to-perception relations.
The formulation of the uncalibrated sensor-based control problem is first presented, then, various computational methods for building adaptive sensorimotor models are derived and analysed. 
The proposed methodology is exemplified with two cases of study: (i) shape control of deformable objects with unknown properties, and (ii) soft manipulation of ultrasonic probes with uncalibrated sensors.
\end{abstract}


\footnotetext{This work is supported in part by the Research Grants Council (RGC) of Hong Kong under grant number 14203917, and in part by PROCORE-France/Hong Kong Joint Research Scheme sponsored by the RGC and the Consulate General of France in Hong Kong under grant F-PolyU503/18.}

\section{INTRODUCTION}
Sensor-based control encompasses a family of methods that exploit feedback information from (typically external) sensors for controlling the robot's motion, and in general, its behaviour. 
On its most fundamental form, it can trace back its origins to the servomechanism problem \cite{Journals:Davison1975}.
Some common examples are visual servoing \cite{Journals:Chaumette2006}, tactile/force servoing \cite{Proceedings:Li2013,Journals:Prats2010}, proximity servoing \cite{Journals:Escaida2015}, aural servoing \cite{Proceedings:Magassouba2016}, deformation/shape servoing \cite{dna2014_ijrr,dna2018_tro}, to name a few cases.

To effectively execute these types of motion tasks, sensor-based controls invariably require some knowledge (at least approximated) of how the robot's motor commands transform into sensory signals. 
This information is captured by the \emph{sensorimotor model} of the robotic system \cite{Journals:Haruno2001}, which besides coordinating action with perception, it can also be used to anticipate the effect that an input motor control will produce on the output sensor measurements \cite{Journals:Pfeifer1997}.
Note, however, that if this information is not known (or is highly uncertain), the robot cannot properly coordinate its motions with sensory feedback.

Existing methods to obtain sensorimotor models require either exact knowledge of its analytical structure \cite{Journals:Wei1998,Journals:Wang2008} (which might not be known) or only compute instantaneous local estimations of it \cite{Proceedings:Hosoda1994} (therefore, they cannot globally describe and control the system).
Compared to these computational approaches, the human brain has a remarkable degree of adaptability that allows it to learn new sensorimotor relations from birth through death and under multiple morphological and perceptual conditions (see e.g. the pioneering study \cite{Journals:Kohler1962}).
Humans can easily coordinate hand motions through a mirror, position unknown tools attached to the body, and even recover (some) mobility after strokes.

Our aim in this paper is precisely to address the design of computational methods that efficiently provide sensor-guided robots with robust adaptation capabilities.
For that, we first formulate the sensorimotor control problem of robots using uncertain perceptual/motor models.
Next, we formulate various structure-based and structure-free methods to adaptively compute these unknown relations.
Finally, the presented methodology is exemplified with two cases of study and discussions about its implementation are given.

The contribution of this work is that it presents a general and intuitive methodology that can be used as a guideline or even a tutorial for researchers working on adaptive sensor-based control of robots with uncertain models. 
The proposed control approach can be used to guide the motion of different robotic platforms (e.g. manipulators, omnidirectional robots, active robot heads) with various sensing modalities (e.g. vision, audio, thermal, attitude, proximity).

The rest of this manuscript is organised as follows: Sec. \ref{sec:preliminaries} presents the preliminaries of the problem; Sec. \ref{sec:model_adaptation} derives different adaptive estimation algorithms; Sec. \ref{sec:cases_of_study} presents the cases of study; Sec. \ref{sec:conclusions} gives final conclusions.

\section{PRELIMINARIES}\label{sec:preliminaries}
\subsection{Notation}
Along this note we use very standard notation.
Column vectors are denoted with bold small letters $\mathbf m$ and matrices with bold capital letters $\mathbf M$. 
Time evolving variables are represented as $\mathbf m_t$, where the subscript $\ast_t$ denotes the discrete time instant, or, the iteration step. 
Gradients of functions $b=\beta(\mathbf m):\mathcal M\mapsto\mathcal B$ are denoted as $\nabla\beta(\mathbf m) = (\partial \beta/\partial\mathbf m{)}^\T$.

\subsection{Control Architecture}
Consider a class of fully-actuated robotic systems whose configuration (e.g. modelling the end-effector pose in a manipulator) is denoted by the vector $\mathbf x_t\in\mathbb R^n$.
In our formulation of the problem, it will be assumed that the motion of robotic system is commanded via a standard position/velocity controlled interface \cite{Journals:Whitney1969,Journals:Siciliano1990} (which is typically found in the large majority of commercial robotic platforms).
With position interfaces, the control commands $\mathbf u_t\in\mathbb R^n$ are given in terms of differential displacement motions as follows:
\begin{equation}
\mathbf x_{t+1} - \mathbf x_t = \mathbf u_t	
\label{eq:displacement_control}
\end{equation}
All methods presented in this note are formulated using the above described position controls, yet, these can be easily transformed into its velocity control equivalent $\mathbf v_t\in\mathbb R^n$ by dividing both sides of \eqref{eq:displacement_control} by the time step $\diff t$ of the servo-loop as 
\begin{equation}
	(\mathbf x_{t+1} - \mathbf x_t)/\diff t = \mathbf u_t/\diff t \approx \mathbf v_t
\end{equation}

\subsection{Configuration Dependant Feedback}
To perform a sensorimotor task, the robot is equipped with a set of $r$ sensors (not necessarily of the same modality) that continuously measure physical quantities whose values depend on the robot's configuration. 
This situation means that relative robot motions produce relative sensory changes.
Some examples of configuration dependent signals (measured using either external or internal/on-board sensors) are: geometric visual features, observed end-effector poses, forces applied onto a surface, proximity to an object, intensity of an audio source, temperature from a heat source, ultrasound images from probe, etc.

The feedback signal from the $i$th sensor is denoted by the vector $\mathbf s_t^i =  g^i(\mathbf x_t)\in\mathbb R^{l_i}$, where the function $ g^i:\mathbb R^n\mapsto\mathbb R^{l_i}$ represents the analytical sensor model that statically relates the instantaneous configuration of the robot with the feedback signal. 
All these signals can be conveniently grouped into a single vector $\mathbf s_t = [{\mathbf s_t^1}^\T,\ldots,{\mathbf s_t^r}^\T{]}^\T= g(\mathbf x_t)\in\mathbb R^l$.
Sensorimotor controls often require to construct a vector of \emph{meaningful} features to quantify and guide the task \cite{Journals:Chaumette2006}.
To this end, let us introduce the (possibly nonlinear) vectorial functional
\begin{equation}
	\mathbf y_t = f(\mathbf s_t) = f( g (\mathbf x_t) ) : \mathbb R^n \mapsto \mathbb R^m
	\label{eq:sensor_model}
\end{equation}
for $m$ as the number of feedback feature coordinates (along this note, we assume that $f(g(\cdot))$ is a smooth functional).
There are three cases with this configuration-dependent structure: $n\ge m$ (more controls than features), $n\le m$ (more features than controls), and $n=m$ (same number of features and controls).
These cases have different properties that determine the controllability of $\mathbf y_t$ by the robot.

\subsection{Sensorimotor Control Problem}
\label{sec:sensorimotor_control}
The differential expression that describes how the motor actions result in changes of feedback features is represented by the first-order difference model:
\begin{equation}
	\mathbf y_{t+1} = \mathbf y_t + \mathbf A_t \mathbf u_t
	\label{eq:differential_model}
\end{equation}
for $\mathbf A_t= [\partial f/\partial \mathbf s_t] [\partial g/\partial \mathbf x_t] \in \mathbb R^{m\times n}$ as the Jacobian matrix of the system (also known as the interaction matrix in the sensor servoing literature \cite{Journals:Cherubini2015}), whose elements depend on the instantaneous configuration $\mathbf x_t$.

The sensorimotor control problem consists in coordinating the motor actions with the feedback signals such that a desired sensory behaviour is achieved.
Without loss of generality, such behaviour is characterised as the set-point regulation of $\mathbf y_t$ towards a constant sensory target $\mathbf y^*$.
The necessary actions $\mathbf u_t$ to approach the target can be computed by minimising the quadratic cost function\footnote{The rational behind the minimisation of \eqref{eq:cost_function} is to find a motor command $\mathbf u_t$ that projects into the sensory space as a vector pointing towards $\mathbf y^*$.}:
\begin{equation}
	J = \left\| \mathbf A_t\mathbf u_t + \lambda\sat(\mathbf y_t-\mathbf y^*) \right\|^2
	\label{eq:cost_function}
\end{equation}
for $\lambda>0$ as feedback gain, and $\sat(\cdot)$ as a vectorial saturation function to ensure that $\mathbf u_t$ satisfies the differential motion condition in \eqref{eq:displacement_control}.
By computing the extremum $\nabla J(\mathbf u_t) = \mathbf 0$, we obtain the \emph{normal equation}
\begin{equation}
	\mathbf A_t^\T \mathbf A_t\mathbf u_t = - \lambda \mathbf A_t^\T \sat(\mathbf y_t-\mathbf y^*)
	\label{eq:normal_equation}
\end{equation}
which exposes the different properties of the three cases relating the relative dimensions of $\mathbf y_t$ and $\mathbf x_t$.

For $n>m$, the solution to the problem can be obtained from \eqref{eq:cost_function} via the right pseudo-inverse of $\mathbf A_t$ as follows \cite{Book:Nakamura1991}:
\begin{equation}
	\mathbf u_t = - \lambda \mathbf A_t^{\T}\left(\mathbf A_t\mathbf A_t^{\T}\right)^{-1} \sat(\mathbf y - \mathbf y^*)
	\label{eq:global_motor_action}
\end{equation}
Note that the above motor action will globally minimise \eqref{eq:cost_function} (i.e. $\|\mathbf y_t- \mathbf y^*\|\rightarrow 0$), as long as the $m$ feature coordinates in $\mathbf y_t$ are linearly independent with respect to $\mathbf x_t$. 
This ensures that the matrix $\mathbf A_t\mathbf A_t^{\T}$ can be inverted.

For $m>n$, the solution is obtained by solving the normal equation \eqref{eq:normal_equation} for $\mathbf u_t$, which yields:
\begin{equation}
	\mathbf u_t = - \lambda \left(\mathbf A_t^\T \mathbf A_t\right)^{-1} \mathbf A_t^{\T} \sat(\mathbf y - \mathbf y^*)
	\label{eq:local_motor_action}
\end{equation}
Substituting \eqref{eq:local_motor_action} into \eqref{eq:cost_function} shows that the cost function can only be locally minimised (i.e. $\|\mathbf y_t-\mathbf y^*\|\rightarrow \eta$, for $\eta > 0$).
The use of redundant features is useful in practice to cope with intermittent feedback from sensors, such as in the case of camera occlusions or malfunctions.

For the trivial case of $n=m$, the matrix $\mathbf A_t$ is square, therefore, the solution is simply obtained via standard matrix inversion $\mathbf u_t = - \lambda \mathbf A_t^{-1} \sat(\mathbf y - \mathbf y^*)$.

\section{CONTINUOUS MODEL ADAPTATION}\label{sec:model_adaptation}

\subsection{Uncertain Sensorimotor Models}
Computing any of the above motor actions requires some knowledge (at least coarse) of the transformation matrix $\mathbf A_t$, which in turn, depends on the sensor and the feature models $g(\cdot)$ and $ f(\cdot)$.
However, if the estimated model is corrupted at some point in time, the robot may no longer properly coordinate action with perception.
This situation may happen when the mechanical structure of the robot is altered (e.g. due to bendings or damage of links) or when the configuration of the perceptual system is changed (e.g. due to relocation of external sensors).

The capability to dynamically estimate sensorimotor models is needed to use robots in many growing (and economically important) fields such as domestic/social robotics, field robotics, autonomous systems, etc, where the sensorimotor conditions are highly uncertain.
Several methods have been proposed to compute or approximate these models (see \cite{Journals:Sigaud2011} for a comprehensive survey on the topic).
In this paper, we coarsely classify these methods into the following two approaches: structure-based estimation and structure-free estimation.
In the following sections, we present the model adaptation problem and provide various solutions to it.

\subsection{Structure-Based Model Adaptation}
These types of algorithms represent calibration-like techniques that aim to estimate the parameters $\boldsymbol\pi\in\mathbb R^p$ of the uncertain model.
Its implementation requires exact knowledge of the analytical structure of the model $\mathbf y_t=f(g(\mathbf x_t)$, which for ease of presentation, we assume it can be linearly parametrisable with respect to $\boldsymbol\pi$ as follows:\footnote{For non-linear model parametrisations, other types of optimisation algorithms (namely non-convex) must be used, whose details are beyond the scope of this expository note.}: 
\begin{equation}
	\mathbf y_t = f(g(\mathbf x_t)) = \mathbf L(\mathbf x_t)\boldsymbol\pi
	\label{eq:linear_parametrisation}
\end{equation}
where $\mathbf L(\mathbf x_t)\in\mathbb R^{m\times p}$ represents a \emph{known} regression-like matrix that captures the properties of the analytical model, and whose elements depend on the configuration vector $\mathbf x_t$.

To compute the vector of estimated parameters $\widehat{\boldsymbol\pi}_t\in\mathbb R^p$, structure-based methods require to first collect a set of $T$ input-output observation points $(\mathbf y_k;\mathbf x_k)$, for $k=1,\ldots,T$ (see e.g. \cite{Journals:Wang2008}). 
Standard methods use this data for computing a quadratic cost function of the following form:
\begin{equation}
	U = \frac{\gamma}{2} \sum_{k=1}^T \left\| \mathbf L(\mathbf x_k)\widehat{\boldsymbol\pi}_t - \mathbf y_k \right\|^2 
\end{equation}
for $\gamma>0$ as a learning gain.
By using \eqref{eq:linear_parametrisation} and after some algebraic operations, it is easy to show that the above function is convex with respect to the estimation error vector $\widehat{\boldsymbol\pi}_t - \boldsymbol\pi$.
Therefore, $U$ can be adaptively minimised with the gradient descent rule\footnote{Along this note, the detailed expressions of function gradients are omitted, yet, these can be easily computed after simple analytical calculations.}:
\begin{equation}
	\widehat{\boldsymbol\pi}_{t+1} = \widehat{\boldsymbol\pi}_t - \nabla U(\widehat{\boldsymbol\pi}_t)
	\label{eq:structure-based_gradient_descent}
\end{equation}
which in the absence of measurement noise and for a sufficient number $T$ of linearly independent observations, it globally minimises the cost $U$ (i.e. $\|\widehat{\boldsymbol\pi}_t - \boldsymbol\pi\|\rightarrow\mathbf 0$, yet, a small estimation error is typically expected in practice).
The adaptive transformation matrix is then simply computed as: 
\begin{equation}
	\widehat{\mathbf A}_t = \frac{\partial}{\partial\mathbf x_t}\left\{\mathbf L(\mathbf x_t)\widehat{\boldsymbol\pi}_t\right\}
\end{equation}

Structure-based approaches have one major disadvantage: its dependency on fixed analytical models. 
Note that since the model's structure \eqref{eq:linear_parametrisation} is explicitly used within the adaptation algorithm \eqref{eq:structure-based_gradient_descent}, these methods are not robust to unknown changes in the mechanical and perceptual conditions.
Furthermore, in many situations, the analytical model might not be available or subject to large uncertainties. 
In the case of advanced robots whose morphology is constantly evolving, fixed analytical models will clearly fail to capture the system's properties.
These issues limit the applicability of structure-based approaches.

\subsection{Structure-Free Model Adaptation}
These types of algorithms have the capability to compute the unknown sensorimotor model in the following manner: (i) entirely from scratch (i.e. without requiring any a-priori knowledge of the model's analytical structure), (ii) on-demand (i.e. they can modify its acquired structure at any time instant so as to identify new relations), and (iii) from data observations only (i.e. by using information from controls and measurements only).

Based on its computation, we coarsely classify these algorithms into the following two general categories: \emph{instantaneous} estimation, and \emph{distributed} estimation.

\paragraph{Instantaneous estimation.} 
As the name suggests, these techniques compute a matrix $\mathbf A_t$ that is only valid at the current (instantaneous) configuration $\mathbf x_t$.
The Broyden rule \cite{Proceedings:Hosoda1994} is one possible example of such technique.
It iteratively computes $\mathbf A_t$ with the following update rule:
\begin{equation}
	\widehat{\mathbf A}_{t} = \widehat{\mathbf A}_{t-1} + \Gamma \frac{\boldsymbol\delta_t - \widehat{\mathbf A}_{t-1} \mathbf u_t}{\mathbf u_t^\T \mathbf u_t} \mathbf u_t^\T
	\label{eq:broyden_rule}
\end{equation}
for $\boldsymbol\delta_t = \mathbf y_{t+1}-\mathbf y_t$, and $0<\Gamma\le 1$ as a tuning gain.
With ``high'' gains $\Gamma\approx 1$, by right-multiplying \eqref{eq:broyden_rule} by $\mathbf u_t$ (namely, projecting the motor action into sensory space), we can see that $\mathbf y_{t+1} \approx \mathbf y_t + \widehat{\mathbf A}_t \mathbf u_t$ is satisfied.
However, using high gains results in a noisy and rapidly changing matrix $\widehat{\mathbf A}_{t}$.
For slow robot motions, the Jacobian matrix is expected to change slowly, therefore, using ``small'' gain values for $\Gamma$ can help to make the computation less responsive, i.e. $\widehat{\mathbf A}_{t}\approx \widehat{\mathbf A}_{t-1}$ as well as to filter out noisy measurements.

Another example of these techniques can be derived from the following (convex) cost function \cite{dna2015_iros}:
\begin{equation}
	V = \frac{\gamma}{2}\left\| \widehat{\mathbf A}_{t} \mathbf u_t - \boldsymbol\delta_t \right\|^2
	\label{eq:cost_function_instantaneous}
\end{equation}
which provides a metric of the accuracy of $\widehat{\mathbf A}_{t}$ at the current configuration $\mathbf x_t$. 
The terms of this unknown matrix are continuously adapted with the rule:
\begin{equation}
	\widehat{a}_{t+1}^{ij} = \widehat{a}_t^{ij} - \nabla V(\widehat{a}_t^{ij})
\end{equation}
where the scalar $\widehat{a}_t^{ij}$ denotes $i$th-row $j$th-column term in $\widehat{\mathbf A}_{t}$.

With instantaneous estimation techniques, the matrix $\mathbf A_t$ must be continuously recalculated with new sensor observations as the robot moves into other configurations.
These types of adaptation techniques do not provide a mechanism for preserving knowledge (i.e. remembering models) of previous configurations.

\paragraph{Distributed estimation.} 
Note that since the feedback feature functional \eqref{eq:sensor_model} is smooth (i.e. differentiable), its Jacobian matrix is expected to smoothly change along the configuration space $\mathbf x_t$.
This means that a \emph{local} Jacobian matrix estimated at a particular configuration point is also valid around the neighbourhood surrounding it.
This simple, yet powerful, idea forms the working principle of distributed estimation techniques.
With these adaptive algorithms, the estimation problem is shared amongst multiple computing units that specialise in a local transformation.

There are many possible implementations of distributed estimation algorithms.
Along this note, we briefly describe a variety based on self-organising maps (see \cite{Book:Kohonen2001} for details).
Consider a network of $N$ computing units spread around the robot's configuration space $\mathbf x_t$.
The following data structure $z^l$ is associated with each computing unit:
\begin{equation}
	z^l = \begin{Bmatrix} \overline{\mathbf x}^l & \widehat{\mathbf A}^l \end{Bmatrix},\quad \text{for}\quad l = 1,\ldots,N
\end{equation}
where $\widehat{\mathbf A}^l$ stands for a local approximation of $\mathbf A_t$ estimated at the configuration point $\overline{\mathbf x}^l$.
There are various methods for establishing the distribution of these units around the robot's workspace, e.g. based on self-organising rules, evenly distributed locations, random point distributions, etc. \cite{Book:Haykin2009}.
For ease of presentation, we assume that the location of these $N$ configuration points $\overline{\mathbf x}^l$ associated with the units has already been established.

Estimation of local transformation matrices is performed by first collecting a data set of $T$ observation points $(\boldsymbol\delta_k;\mathbf u_k)$, for $k=1,\ldots,T$.
Then, the following local cost function for the $l$th unit is computed:
\begin{equation}
	W^l = \frac{\gamma}{2} \sum_{j\in\mathcal B} h^{lj} \left\| \widehat{\mathbf A}^l \mathbf u_j - \boldsymbol\delta_j \right\|^2
	\label{eq:cost_function_distributed}
\end{equation}
for $\mathcal B$ as a local ball centred at the $l$th unit, and $h^{lj}$ as its Gaussian neighbourhood function computed as
\begin{equation}
	h^{lj} = \exp \left( - \frac{ \|\overline{\mathbf x}^l - {\mathbf x}^j \|^2 }{2\sigma^2} \right)
\end{equation}
where $\sigma>0$ determines the ball's radius.
In this method, the idea is to make use of neighbouring data (whose contribution decreases with its distance to the centre unit $l$) for approximating the local transformation matrix.
The update rule to adaptively compute the $i$th row $j$th column term of $\widehat{\mathbf A}^l_t$ is as follows:
\begin{equation}
	\widehat{a}_{t+1}^{l,ij} = \widehat{a}_t^{l,ij} - \nabla W(\widehat{a}_t^{l,ij})
\end{equation}

Once the $N$ local cost functions \eqref{eq:cost_function_distributed} have been minimised, the network is trained to perform sensorimotor transformations with each of its units.
In order to implement motor commands as the ones derived in Sec. \ref{sec:sensorimotor_control}, a local transformation matrix that best matches the current position $\mathbf x_t$ must be retrieved from the $l$ units.
This can be done by solving the search problem
\begin{equation}
l = \argmin_j \left\{ \|\overline{\mathbf x}^j - \mathbf x_t\| \right\}
\label{eq:winning_node}
\end{equation}

Note that this adaptation approach can be combined with the previous instantaneous estimation technique (or possibly others) by defining a cost function $H^l = V + W^l$ that exploits the current feedback measurements. 
This allows to also quantify the accuracy of the model based on new sensory data; the cost function can then be minimised with similar gradient descent tools as before.

Compared to the previous estimation approaches, distributed estimation requires more data to approximate the sensorimotor model. 
However, these methods can effectively preserve local knowledge within its computing units and does not require any prior analytical representation of the robot's action-to-perception relations.


\section{CASES OF STUDY}\label{sec:cases_of_study}
\begin{figure}[t]
				\centering
				\includegraphics[width = \columnwidth]{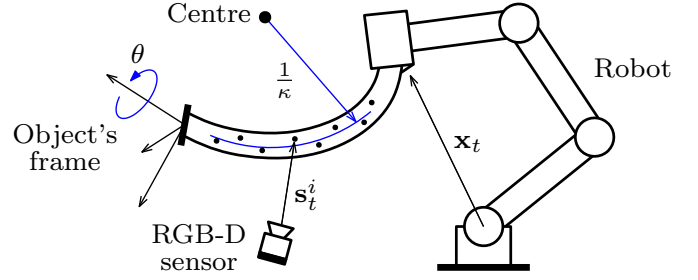}
				\caption{Shape servoing system composed of a 4-DOF manipulator, an elastic object, and a 3D vision sensor.}
				\label{fig:shape_servoing}
\end{figure}

\subsection{Visual Shape Servoing of Elastic Objects}
To exemplify our methodology, consider the setup depicted in Fig. \ref{fig:shape_servoing}, where a 3D camera captures point clouds of a beam-like elastic object manipulated by a robot.
Let us denote the captured 3D points by $\mathbf s^i_t$, for $i=1,\ldots,r$ (note that the number $r$ is generally in the order of hundreds).
The task to be performed is to automatically deform the object into a desired shape.
We can use the point cloud to approximate a the object's backbone (represented as the blue curve in Fig. \ref{fig:shape_servoing}).
With this geometric information, we compute the feature vector defined as follows:
\begin{equation}
	\mathbf y_t = 
	\begin{bmatrix}
	\kappa & \theta
	\end{bmatrix}^\T
\end{equation}
for $\kappa>0$ as the object's curvature, and $\theta$ as the angle of the object's bending with respect to it's frame, see \cite{dna2019_front} for details.
For this task, \emph{model adaptation} can be performed with distributed estimation algorithms.
These approaches provide an efficient solution to the highly nonlinear transformation problem of relating robot poses to object deformations (note that deformation models are hard to compute analytically).
For that, several computing units must be first defined at key end-effector poses, e.g. ranging from fully stretched to varying bending configurations; local sensory observations can then be collected for approximating the model.
Since $n>m$, the motor command is computed as in \eqref{eq:global_motor_action}, with a target shape defined as $\mathbf y^* = [\kappa^*,\theta^*{]}^\T$.

\subsection{Multi-Modal Scanning with Ultrasound Probes}
\begin{figure}[t]
				\centering
				\includegraphics[width = \columnwidth]{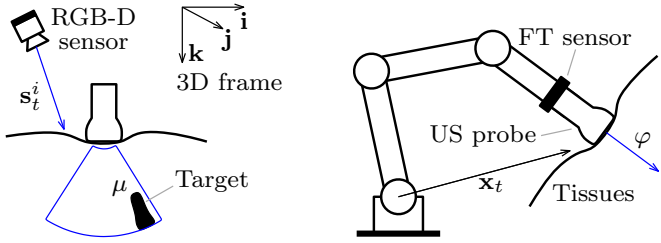}
				\caption{Multi-modal system composed of a robot manipulator, an ultrasound probe, a force sensor and a 3D camera.}
				\label{fig:ultrasound_probe}
\end{figure}

Consider the setup in Fig. \ref{fig:ultrasound_probe}, which depicts a robot performing automatic scanning of tissues with an ultrasound probe \cite{maria2019_crc} (we assume the robot has 6-DOF).
This system is instrumented with a force/torque sensor and a 3D camera.
Let us denote by $\varphi$ the (normal) force applied onto tissues, by $\mu$ the image location of the ultrasound feature of interest, and by $\boldsymbol\omega$ the probe's 3D orientation.
The task to be performed is conveniently described with respect to the body's 3D frame.
It consists in positioning $\mu$ at the centre of the ultrasound image, while applying a desired normal force and controlling the probe's pose over the tissues.
Note that this relative orientation can be computed from the 3D point clouds $\boldsymbol\omega = q(\mathbf s_t^1,\ldots,\mathbf s_t^r)$.
The task's feature vector is defined as
\begin{equation}
	\mathbf y_t =
	\begin{bmatrix}
	\mu & \varphi & \boldsymbol\omega
	\end{bmatrix}^\T
\end{equation}

The models for the above feature coordinates are simple to analytically derive, namely using Hooke's law for $\varphi$, horizontal image displacements for $\mu$, and affine/homogeneous transformations for $\boldsymbol\omega$.
Therefore, \emph{model adaptation} can be performed using structure-based algorithms as in \eqref{eq:linear_parametrisation}.
With these approaches, we can robustify the sensor-guided task by continuously calculating unknown task parameters such as: stiffness of soft tissues, relative location of ultrasound features, and robot-camera-body transformations.
Since $n>m$, the motor command is also computed as in \eqref{eq:global_motor_action}, with a set-point feature defined as $\mathbf y^* = [\mu^*, \varphi^*, \boldsymbol\omega^*{]}^\T$.

\section{CONCLUSIONS}\label{sec:conclusions}

In this expository paper, we addressed the problem of computing adaptive sensorimotor models for robots with uncalibrated sensory feedback and/or uncertain morphology.
A general sensor servoing approach was first formulated using an energy minimisation approach.
Then, we derived various methods for providing these controllers with continuous model adaptation capabilities.
Two cases of study we presented to illustrate the proposed methodology.

The presented sensorimotor controls are formulated based on the assumption that the feedback signals dependent on the robot's configuration only.
Although this condition can be fairly used to represent many sensor-guided applications, it may not be the most accurate model for describing tasks where the measurements also depend additional variables (e.g. manipulating fabrics with infinite dimensional configurations) or even time-varying states (e.g. controlling the effect of cosmetic lasers stimulating skin tissues).
The development of more general sensor models is still an open research problem.

The presented model adaptation methods allow robots to perform sensor-guided tasks even when its sensorimotor model is not known or might suddenly change.
For example, robots can adapt to unknown sensor configurations and/or morphologies.
By understanding the principle of how sensors models can be effectively created from scratch and adapted on-the-fly, we hope to build machines with more resilient properties that allow them to perform long-term operations with minimal supervision.
These advanced capabilities are needed to advance towards building truly autonomous robots.

\balance

\bibliography{david.bib,bibliography.bib} 
\bibliographystyle{IEEEtran}

\end{document}